\documentclass[conference]{IEEEtran}
\IEEEoverridecommandlockouts
\usepackage{cite}
\usepackage{float}
\restylefloat{table}
\usepackage{amsmath,amssymb,amsfonts}
\usepackage{algorithmic}
\usepackage{graphicx}
\usepackage{textcomp}
\usepackage{xcolor}
\usepackage{xspace}
\usepackage{comment}
\usepackage{subfigure}
\floatstyle{plaintop}
\restylefloat{table}
\usepackage[T1]{fontenc}
\usepackage{pgfplots}
\def\BibTeX{{\rm B\kern-.05em{\sc i\kern-.025em b}\kern-.08em T\kern-.1667em\lower.7ex\hbox{E}\kern-.125emX}}

\begin{document}
\newcommand{\jina}{{ContrastNER}\xspace}
\newcommand{\DataCloud}{{DataCloud}\xspace}
\title{\jina: Contrastive-based Prompt Tuning for Few-shot NER
%
%
%
%
%

\thanks{The work in this paper was partially funded by the projects DataCloud (H2020 101016835), enRichMyData (HE 101070284), Graph-Massivizer (HE 101093202), UPCAST (HE 101093216), and BigDataMine (NFR 309691). The experiments were enabled by resources provided by the Swedish National Infrastructure for Computing (SNIC) at Chalmers Centre for Computational Science and Engineering (C3SE) partially funded by the Swedish Research Council through grant agreement no. 2018-05973.}
}

\author{\IEEEauthorblockN{Amirhossein Layegh$^{\dag}$, Amir H. Payberah$^{\dag}$, Ahmet Soylu$^{\ddag}$, Dumitru Roman$^{\S}$, Mihhail Matskin $^{\dag}$}
\IEEEauthorblockA{
$^{\dag}$KTH Royal Institute of Technology, Sweden
$^{\ddag}$Oslo Metropolitan University, Norway
$^{\S}$SINTEF AS, Norway\\
$^{\dag}$\{amlk, payberah, misha\}@kth.se 
$^{\ddag}$ahmet.soylu@oslomet.no
$^{\S}$dumitru.roman@sintef.no
}}

\maketitle

\begin{abstract} \label{abstract}
Prompt-based language models have produced encouraging results in numerous applications, including Named Entity Recognition (NER) tasks. NER aims to identify entities in a sentence and provide their types. However, the strong performance of most available NER approaches is heavily dependent on the design of discrete prompts and a verbalizer to map the model-predicted outputs to entity categories, which are complicated undertakings. To address these challenges, we present \jina, a prompt-based NER framework that employs both discrete and continuous tokens in prompts and uses a contrastive learning approach to learn the continuous prompts and forecast entity types. The experimental results demonstrate that \jina obtains competitive performance to the state-of-the-art NER methods in high-resource settings and outperforms the state-of-the-art models in low-resource circumstances without requiring extensive manual prompt engineering and verbalizer design.
\end{abstract}


\section{Introduction} \label{introduction}
{\em Named Entity Recognition (NER)} aims to recognize and classify named entities, such as person and location, into the appropriate concept classes. NER plays a crucial role in various applications, including information extraction, ontology population, question answering, machine translation, and semantic annotation, to name a few. Despite extensive research in this area, the state-of-the-art solutions still need more generalization and extensibility due to their reliance on domain-specific knowledge resources such as annotated training corpus. Considering that resources for data annotation are scarce in many domains and annotating a large corpus of text labeled in some domains requires the expertise of experts, low-resource NER tasks become a complex problem.

\begin{figure}
    \centering
    \resizebox{.5\textwidth}{!}{
    \includegraphics{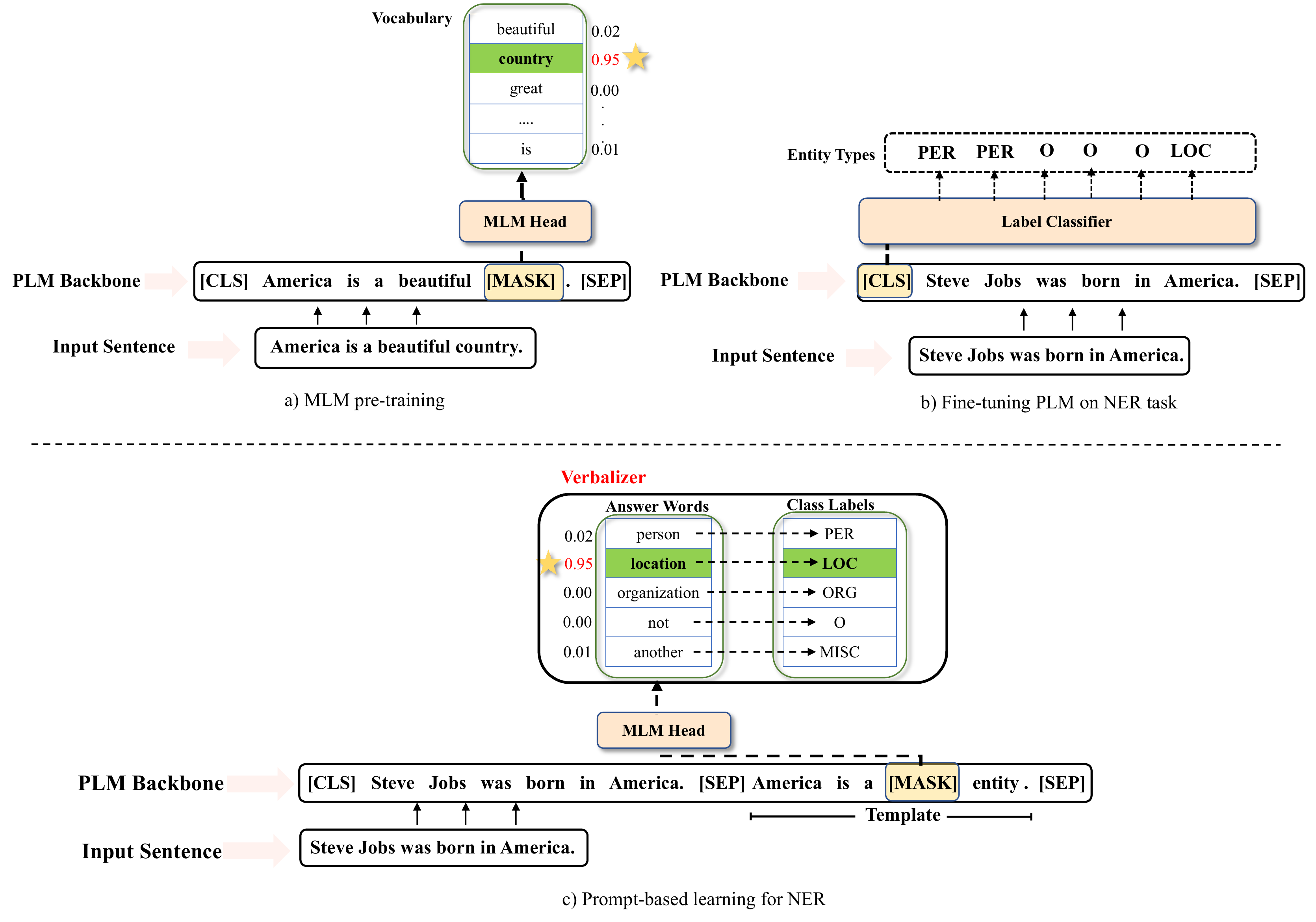}}
    \caption{Illustration of two different paradigms for solving NER task. The top image shows pre-training a language model using the MLM objective~(\ref{fig:introduction:NER}a) and fine-tuning it for a NER task to predict the entity type for each word in an input sentence~(\ref{fig:introduction:NER}b). The bottom image shows prompting the input sentence with a template to transform the NER task into an MLM problem to predict the type of a candidate entity in the input sentence~(\ref{fig:introduction:NER}c).}
    \label{fig:introduction:NER}
\end{figure}

Pre-training a model on a rich-resource dataset and fine-tuning it on a low-resource downstream task is becoming more prevalent~\cite{raffel2020exploring}. Typically, in the context of NER, the pre-trained step includes training a model using Masked Language Modeling (MLM) to predict the probability of the observed textual data. Then, the fine-tuning stage fine-tunes the Pre-trained Language Model (PLM) developed in the preceding step to predict the type of identified entities (Figure~\ref{fig:introduction:NER}(a) and \ref{fig:introduction:NER}(b)). Recent NER models based on this methodology have demonstrated exemplary performance on NER tasks~\cite{devlin2018bert,yamada2020luke}. 
However, to ensure high-quality learning, the downstream task requires a substantial amount of labeled data for satisfactory performance. Thus, if no annotated resources exist in the target domain, the model cannot identify the corresponding entity types, indicating poor generalization in low-resource circumstances. 

Existing pre-training and fine-tuning NER models have an additional obstacle in the presence of two distinct pre-training and fine-tuning objectives (i.e., MLM in the pre-training and predicting the entity types in the fine-tuning). Initiated by GPT-3~\cite{brown2020gpt}, a \textit{prompt-based} technique is one solution to bridging the gap between two different objectives. In this approach, the objectives of both the pre-training and fine-tuning stages are formulated as an MLM problem where the model directly predicts a textual answer to a given prompt~\cite{gao2020making,liu2021gpt}. The generated answer will then be mapped to a class label using a \textit{verbalizer}~\cite{schick2020exploiting} (Figure \ref{fig:introduction:NER}(c)). A recent body of work investigates the setting of low-resource NER by applying prompt-based approaches to integrate the objectives of pre-training and fine-tuning phases~\cite{lewis2019bart,chen2021lightner,liu2022qaner}. However, these solutions depend significantly on expensive discrete prompt engineering.

To address these challenges, we present \jina for the NER task that leverages the few-shot learning capabilities of PLMs without manual prompt engineering and verbalizer design. To this end, we introduce {\em soft-hard prompt tuning} for automatic prompt search in a mixed space of continuous and discrete prompts. Moreover, we employ contrastive learning to combine learning the soft-hard prompt and predicting the entity type without explicitly designing a verbalizer. 

In summary, we present the following contributions:
\begin{itemize} 
\item We propose \jina, a prompt-based model for NER learning using a low-resource dataset and without manual search for appropriate prompts and creation verbalizers. To the best of our knowledge, our work is the first model that uses soft-hard prompt tuning and contrastive learning for prompt-based NER learning.
\item We conduct experiments on four publicly available NER datasets demonstrating that our method outperforms the state-of-the-art low-resource prompt-based NER learning techniques.
\end{itemize}

\section{Preliminaries and Problem Definition} \label{preliminaries}
In this section, first, we introduce the basic concepts of NER and then explain the fine-tuning of standard NER and prompt-based learning for few-shot NER.

\subsection{Named Entity Recognition (NER)} \label{sec:preliminaries:NER}
NER aims to identify \textit{entities} in a sentence and provide their \textit{entity types}. Entities (e.g., noun/verb phrases) are the main parts of a sentence, and entity types are the labels for each entity, which are characterized based on predefined categories, such as place, person, organization, and suchlike.

In a NER dataset, an example typically is a pair of $(\textbf{X}, \textbf{Y})$, where $\textbf{X}$ = \{$x_1, \cdots, x_m$\}, is a sentence containing $m$ words, and $\textbf{Y}$ = \{$y_1, \cdots, y_m$\} are the corresponding labels for each word in $\textbf{X}$ that specify their entity types. For instance, the labels for $\textbf{X}$ = {\tt\{"Steve", "Jobs", "was", "born", "in", "America", "."\}}, are $\textbf{Y}$ = {\tt\{"PERSON", "PERSON", "O", "O", "O", "LOCATION", "O"\}}, where {\tt"PERSON"}, {\tt"O"}, and {\tt"LOCATION"} indicate the organization entity type, not a named entity, and location entity type, respectively. The goal of a NER task is to predict the label (entity type) $y_i$ for each word $x_i$. There are various methods for creating a NER task, which are discussed in the following sub-sections.

\subsection{Pre-training and Fine-tuning Models for NER} \label{preliminaries:pre-training}
One approach to training a model for a NER task involves fine-tuning PLM on a downstream NER task. This procedure usually entails two steps: (1) \textit{pre-training} and (2) \textit{fine-tuning}. In the first step, a model is trained using massive unlabeled text data. To this end (as illustrated in Figure~\ref{fig:introduction:NER}(a)), a certain percentage of the input tokens (words) are randomly corrupted (replaced with {\tt [MASK]} token), and the final corresponding vectors are given into a softmax over the vocabulary to predict these masked tokens. This process is known as \textit{Masked Language Modeling} (MLM). Finally, the checkpoint of the trained model, $\mathcal{L}$, is saved as a PLM and will be fine-tuned on different downstream tasks.

In the second stage, using a NER dataset, the PLM $\mathcal{L}$ is fine-tuned on a downstream NER task. To do this, each input sentence $\textbf{X} = \{x_1, \cdots, x_m\}$ in the dataset is first converted into the input sequence as \{{\tt [CLS]}, $x_1, \cdots, x_m$, {\tt [SEP]}\}, where {\tt [CLS]} is a special token and {\tt [SEP]} is the end of the sequence. Then $\mathcal{L}$ encodes all tokens of $\textbf{X}$ into an input embedding \(\{ h_{{\tt[CLS]}}, h_{x_1}, h_{x_2}, \cdots, h_{x_m}, h_{{\tt [SEP]}}\}\). Typically, a label-specific classifier is employed to compute the probability distribution of $h_{{\tt[CLS]}}$ over the entity label space {\tt \{"PER", "ORG", "MISC", $\cdots$\}} to assign correct entity types (Figure~\ref{fig:introduction:NER}(b)). Finally, $\mathcal{L}$ is fine-tuned by minimizing the cross-entropy of the loss function.

\subsection{Prompt-based Learning for Few-shot NER}\label{preliminaries:prompt}
A rich-resource dataset is usually required to train a NER model, which is not always available. Therefore, we should examine \textit{few-shot NER}, where very few examples of each label (entity type) in the dataset are available. A common strategy for few-shot NER is to adapt a PLM $\mathcal{L}$ trained with MLM on a rich-resource dataset to a low-resource target NER dataset containing few examples per entity type~\cite{cui2021template,chen2021lightner,ziyadi2020example}.

One method to train a model for few-shot NER is to use prompt-based learning~\cite{liu2021pre}. This approach reformulates the downstream NER task as an MLM problem using a textual prompt template. To this end, it is necessary to define two functions: (1) a prompt template \(\mathcal{T}\) and (2) a verbalizer $\mathcal{M}$. The prompt template $\mathcal{T}$ organizes the input sentence $\textbf{X}$, masked token, and prompt tokens as $\mathcal{T}(\textbf{X})$ = \{$\textbf{X}$ {$\langle${\tt candidate\_entity}$\rangle$ {\tt is a [MASK] entity}}\} (see Figure~\ref{fig:introduction:NER}(c)). For instance, in a NER task with $\textbf{X}$ = \{{\tt Steve Jobs was born in America.}\}, the $\mathcal{T}(\textbf{X})$ = {\tt \{Steve Jobs was born in America. America is a [MASK] entity\}}, where {\tt America} represents $\langle${\tt candidate\_entity}$\rangle$.

After constructing $\mathcal{T}(\textbf{X})$ for each entity (e.g., {\tt Steve} and {\tt America}), the PLM $\mathcal{L}$ will predict the mask token {\tt [MASK]}, which will be translated into an entity type by a verbalizer $\mathcal{M}$. For example, a verbalizer $\mathcal{M}$ can be defined as below:
\\
\\
\begin{center}
    $\mathcal{M}$({\tt "organization"}) $\rightarrow$ {\tt ORG}\\ 
    $\mathcal{M}$({\tt "person"}) $\rightarrow$ {\tt PER}\\ 
    $\mathcal{M}$({\tt "location"}) $\rightarrow$ {\tt LOC}\\
    $\cdots$\\
\end{center}

Handcrafting prompt templates using discrete tokens in natural language (e.g., "$\langle${\tt candidate\_entity}$\rangle$ {\tt is}") is challenging. An alternative approach is to use {\em soft prompts}, where continuous tokens are added to the input and are updated rather than discrete tokens~\cite{lester2021power}. P-tuning~\cite{liu2021gpt} is a model that uses prompt tuning to prevent prompt engineering with discrete tokens. P-tuning applies $\mathcal{T}$ on an input $\textbf{X}$ and creates \{$\textbf{X}$ $h_0 \cdots h_n$ {\tt [MASK]}\}, where $\{h_0 \cdots h_n\}$ are continues prompts. Consequently, P-tuning uses PLM $\mathcal{L}$ to create an embedding of {\tt[MASK]} and applies a lightweight neural network, called {\em prompt encoder}, to learn the embedding of continuous prompts. During fine-tuning, $\mathcal{L}$'s parameters are frozen, and only the parameters of the prompt encoder and $\{h_0 \cdots h_n\}$ are updated.~\cite{liu2021gpt}. 

For example, for the input sentence $\textbf{X}$ = {\tt \{Steve Jobs was born in America.\}} and the trainable continuous prompt $\{h_0 \cdots h_n\}$, the prompt template is $\mathcal{T}(\textbf{X})$ = {\tt \{Steve Jobs was born in America. $h_0 \cdots h_i$ [MASK]\}}. During fine-tuning, $\mathcal{L}$ predicts the {\tt [MASK]} embedding. They use sequence tagging to solve NER tasks by assigning labels marking at the beginning and the end of some entity classes, and the prompt encoder learns the embedding of $\{h_0 \cdots h_n\}$.

\section{Related work} \label{relatedwork}
\subsection{Named Entity Recognition} \label{relatedwork:NER}
Using sequence labeling and Conditional Random Fields (CRF)~\cite{lafferty2001conditional} to associate each word in the input text with a label is a popular method to formulate NER tasks. Earlier works investigated utilizing several neural architectures such as BiLSTM~\cite{sepp1997bilstm} or CNN~\cite{kwak2016cnn} and training the model in a supervised learning paradigm, which often involves a large amount of annotated data~\cite{mahovy2016end}\cite{chiu2016named}\cite{chen2019grn}\cite{ilic2018deep}. Recently, PLMs have shown significant improvements in NER by using large-scale transformer-based architectures as the backbone for learning text representation~\cite{devlin2018bert}. The state-of-the-art results achieved by~\cite{yamada2020luke,luo2020hierarchical} propose a new pre-trained contextualized representation of words with a transformer-based architecture pre-trained on a large set of the entity-annotated corpus. Despite the satisfactory performance achieved by PLMs in NER tasks, these approaches are primarily developed for supervised rich-resource NER datasets, which have limited generalization capability in low-resource datasets~\cite{fritzler2019few}.

A more recent line of work has focused on enhancing model learning capabilities to maximize the use of existing sparse data and reduce reliance on data examples. One line of earlier work on low-resource NER has included prototype-based techniques that apply meta-learning~\cite{ravi2016optimization} to few-shot NER. The majority of these approaches~\cite{li2020metaner,ziyadi2020example,ding2021few} employ a prototype-based metric, often k-nearest neighbor, to learn the representation of similar entities from different domains. However, in these approaches, the network parameters of the NER model cannot be updated, resulting in poor performance when adapting the model to a target domain with few available examples.

\subsection{Prompt-based Learning} \label{relatedwork:prompt_based_PLMs}
Recently, starting from GPT-3~\cite{brown2020gpt}, prompt-based learning has arisen to bridge the gap between the objectives of pre-training and fine-tuning. These approaches reformulate the downstream task by incorporating a template that transforms the input sentence into one that resembles the examples solved during pre-training. This strategy aims to fully apply acquired knowledge from pre-training to the downstream task. As stated in~\cite{scao2021many}, a well-chosen prompt can be equivalent to hundreds of data points; hence, prompt-based learning can be highly advantageous for solving low-resource tasks. Another line of research studied a lightweight alternative to fine-tuning known as prompt-tuning, which optimizes a continuous task-specific vector as a prompt while leaving the parameters of the language model unchanged~\cite{li2021prefix}\cite{liu2021gpt}. However, the efficiency of prompt-tuning for complex sequence labeling tasks such as NER has yet to be verified.

Regarding NER, TemplateNER~\cite{cui2021template} is a template-based prompt method using BART~\cite{lewis2019bart} that treats the NER task as a language model ranking problem. This model manually creates a template for each class and separately populates each created template with all candidate entity spans extracted from the input sentence. The model then assigns a label to each entity candidate span based on the respective template score. In contrast to TemplateNER, instead of manually searching for an appropriate template, which is labor-intensive and time-consuming, we propose inserting some adjustable tokens into the template to search automatically for the ideal prompt template. LightNER~\cite{chen2021lightner} introduces prompt-tuning to the attention layer by incorporating continuous prompts into the attention layer. Moreover, LightNER constructs a unified semantic aware space to remove label-specific classifiers placed on top of encoders. Like LightNER, we eliminate the label-specific layers that map the generated answer to a class label. Instead of the attention layer, we insert soft prompts into the input sentence, and the creation of an answer space is no longer required.

\subsection{Contrastive Learning} \label{relatedwork:contrastive_learning}
Recent works have studied contrastive learning for visual representation~\cite{chen2020simple}\cite{le2020contrastive}, graph representations~\cite{velickovic2019deep}, and a variety of NLP tasks including sentence-level text representation~\cite{gao2021simcse}\cite{iter2020pretraining}, relation extraction~\cite{ding2021prototypical}, machine translation~\cite{yang2019reducing}, sentiment analysis~\cite{xu2022making}, knowledge graph embeddings~\cite{bose2018adversarial}, caption generation~\cite{vedantam2017context}. Contrastive representation learning is intuitively similar to learning by comparison in that it aims to project similar samples close together in the embedding space while mapping dissimilar samples further apart~\cite{hadsel2006contrastive}. Khosla et al.~\cite{khosla2020supervised} study applying contrastive learning in a fully supervised setting and demonstrated that batch contrastive techniques outperform the cross-entropy loss and traditional contrastive losses, such as triplet~\cite{schroff2015facenet}, max-margin~\cite{liu2016large}, and the N-pairs~\cite{sohn2016improved} loss. This work applies supervised batch contrastive learning to prompt-based few-shot NER to differentiate between different classes (entity types) in a sentence.

\begin{figure*}[t]
    \centering
    \scalebox{0.75}{
    \includegraphics[width=\textwidth]{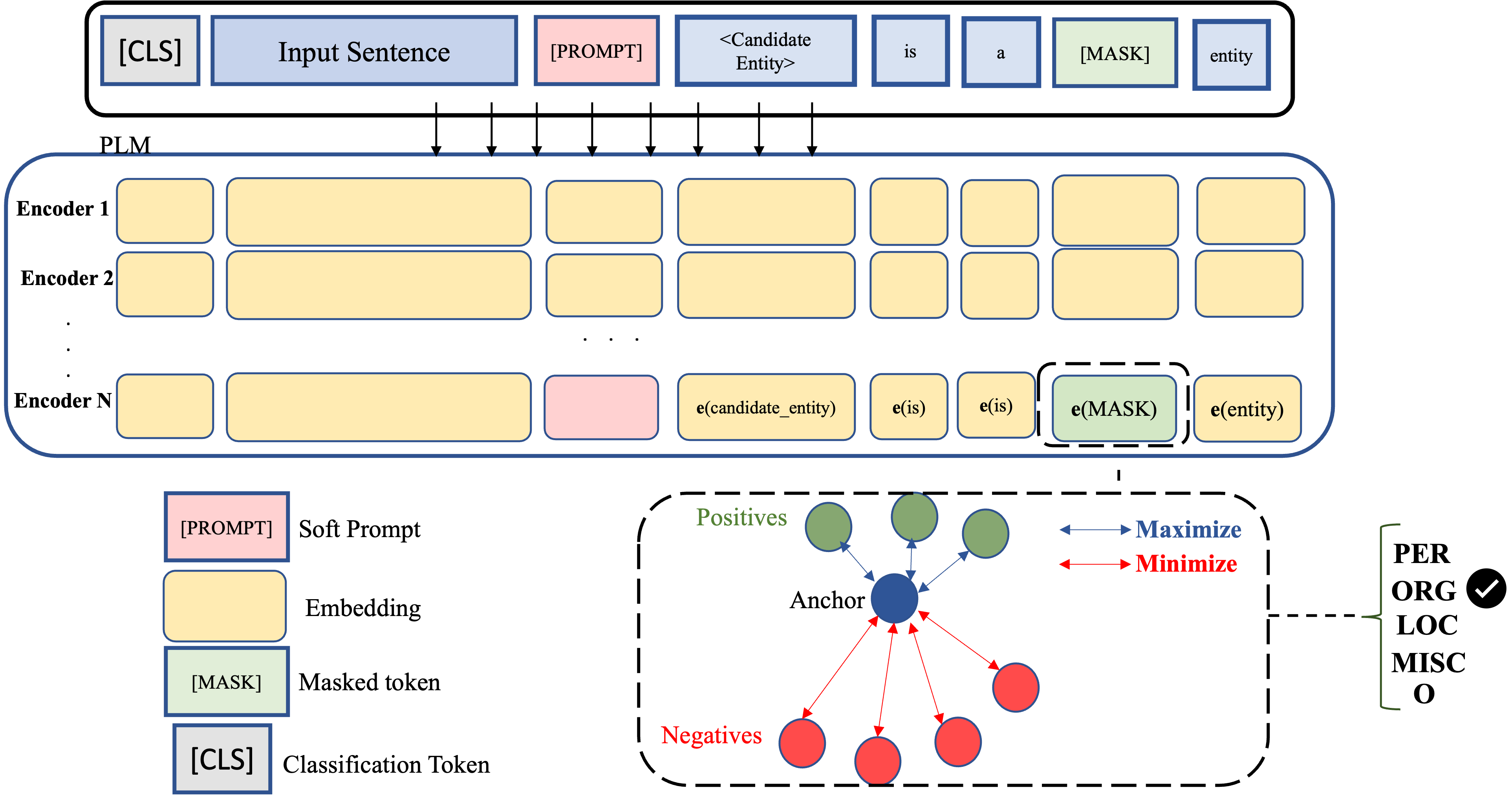}\hspace{0.5em}}
    \caption{Overview of \jina.}
    \label{solution:overview}
\end{figure*}

\section{Overview of \jina} \label{solution}
In this section, we introduce \jina, our proposed prompt-based model for few-shot NER, using two main techniques: (1) soft-hard prompt tuning (\ref{solution:soft_hard}) and (2) a verbalizer-free mechanism (\ref{solution:verbalizer-free}). In the following subsections, we explain the details of \jina (depicted in Figure~\ref{solution:overview}). 

\subsection{Soft-Hard Prompt Tuning} \label{solution:soft_hard}
Finding appropriate discrete prompt (a.k.a. {\em hard prompt})\footnote{Throughout the paper, we will interchangeably use {\em discrete prompt} and {\em hard prompt} but convey the same meaning as discrete prompt templates.} templates in natural language for NER models is challenging~\cite{jiang2020can,reynolds2021prompt,gao2020making,cui2021template}. Existing hard prompt-based NER models, such as LAMA~\cite{petroni2019language}, have shown that a single word change in prompts can cause an extreme difference in the results; hence an approach that is not sensitive to different discrete prompts would be advantageous in prompt-based learning models. As explained in Section~\ref{preliminaries:prompt}, continuous prompts (a.k.a. {\em soft prompts}) and tuning them using a prompt encoder is an approach to solve the hard prompt challenges~\cite{li2021prefix}\cite{liu2021gpt}\cite{lester2021power}. However, this approach leads to two challenges: (1) it is still inferior to fine-tuning approach when the model size is not significant~\cite{liu2021p}, and (2) adding a prompt encoder to learn the soft prompt results in learning extra parameters related to the prompt encoder.

To address these challenges, we propose the {\em soft-hard prompt}, where we use both soft (continuous) and hard (discrete) prompts in the fine-tuning paradigm. To this end, we transform a given input sentence $\textbf{X}$ = \{$x_1, \cdots, x_m$\}, into $m$ new input sentences using a prompt template $\mathcal{T}$, such that each new sentence has three parts: (1) the original sentence $\textbf{X}$, (2) a soft prompt \{$h_0 \cdots h_n$\}, and (3) a masked hard prompt. Suggested by~\cite{cui2021template}, the masked hard prompt initialized as "$\langle${\tt candidate\_entity}$\rangle$ {\tt is a [MASK] entity}". For example, $\textbf{X}$ = \{{\tt "Steve", "Jobs", "was", "born", "in", "America"}\} will be transformed into six sentences as shown in Table~\ref{solution:table:examples}. To learn the hard token {\tt [MASK]} and soft tokens \{$h_0 \cdots h_n$\}, we use a PLM $\mathcal{L}$ and pass $\mathcal{T}$(\textbf{$X$}) as input to it. Then we optimize the soft tokens by the loss function $L_S$ (Equation~\ref{sol:cross}) computed as the cross-entropy between the actual entity type and the predicted {\tt [MASK]} entity type by $\mathcal{L}$ (the details are in Section~\ref{solution:verbalizer-free}).

\begin{table}[t] 
\caption{\footnotesize A sentence \textbf{X} and generated soft-hard prompts by $\mathcal{T}(\textbf{X})$.} \label{solution:table:examples}
\resizebox{.5\textwidth}{!}{%
\begin{tabular}{|l|}
\hline
\multicolumn{1}{|c|}{\textbf{A sample input sentence $\textbf{X}$}} \\ \hline
{\tt Steve Jobs was born in America}\\ \hline\hline
\multicolumn{1}{|c|}{\textbf{Generated soft-hard prompts by $\mathcal{T}(\textbf{X})$}}\\ \hline
\tt Steve Jobs was born in America $[h_1] \cdots [h_p]$ Steve is a [MASK] entity.\\\tt Steve Jobs was born in America $[h_1] \cdots [h_p]$ Jobs is a [MASK] entity.\\\tt Steve Jobs was born in America $[h_1] \cdots [h_p]$ was is a [MASK] entity.\\\tt Steve Jobs was born in America $[h_1] \cdots [h_p]$ born is a [MASK] entity.\\\tt Steve Jobs was born in America $[h_1] \cdots [h_p]$ in is a [MASK] entity.\\\tt Steve Jobs was born in America $[h_1] \cdots [h_p]$ America is a [MASK] entity.\\ \hline
\end{tabular}%
}
\end{table}
\subsection{Verbalizer-Free Mapping} \label{solution:verbalizer-free}
Previous prompt-based approaches for NER often require a verbalizer to create a one-to-one mapping between the predicted token for {\tt [MASK]} in a template and an entity type~\cite{liu2021pre}, which is computationally intensive. In addition, using a manually crafted verbalizer in few-shot NER, where the label space of the source and target domains may differ, leads to incompatibilities that negatively impact the model generalization~\cite{schick2020s}. Moreover, a verbalizer usually only considers the semantic relationship between the predicted token and a few words specified in the verbalizer (e.g., the $location$ for {\tt LOCATION} entity and the $person$ for {\tt PERSON} entity). However, the predicted token may have semantic relationships with other words (e.g., $place$ for {\tt LOCATION} entity, and $people$ for {\tt PERSON}), which the verbalizer will ignore.

To tackle the challenges of verbalizers mentioned above, we unify the models for learning the soft-hard prompt and predicting the entity type and propose a novel objective that includes supervised contrastive learning terms for fine-tuning a pre-trained language model. To do so, instead of forwarding the embedding of {\tt [MASK]} through a verbalizer to detect the label (entity type), we use PLM $\mathcal{L}$ to directly predict the embedding of the label by fine-tuning it using a contrastive learning-based task~\cite{xu2022making,hadsel2006contrastive}.

During training $\mathcal{L}$, we first transform all input examples within the training batch $\mathcal{B}$ using the soft-hard prompt template $\mathcal{T}$. Then, for each example $i \in \mathcal {B}$ (with label $y_i$), we select a set of {\em positive} examples with similar labels to $i$, \(P(X) = \{i^+\ | y_{i^+}=y_i, {i^+} \in \mathcal{B}, {i^+}\ne i\}\) and a set of {\em negative} examples with different labels, \(N(i) = \{i^-\ | y_{i^-} \ne y_i, i^- \in \mathcal{B}, i^-\ne i\}\). Assume $t_i$ denotes the embedding of the predicted label of {\tt [MASK]} for the example $i$, and $t_{i^+}$ and $t_{i^-}$ are the embedding of the labels of positive and each negative examples, respectively, where $i^+ \in P(i)$ and $i^- \in N(i)$. Then, we use the contrastive model~\cite{gao2021simcse} to maximize the {\em within-class} similarity $sim(t_i, t_{i^+})$ of $t_i$ and $t_{i^+}$, and minimize the {\em between-class} similarity $sim(t_i, t_{i^-})$ of $t_i$ and $t_{i^-}$, where $sim(t_1, t_2)$ is the cosine similarity of $t_1$ and $t_2$. We define the contrastive learning loss function as below:
\begin{equation}
L_C = -\log \sum_{i^+ \in P(i)}^{} \frac{e^{sim(t_i, t_{i^+})/\tau}}{e^{sim(t_i, t_{i^+})} + e^{sim(t_i, t_{i^-})/\tau}}
\end{equation} 
where $\tau$ is a temperature hyperparameter. In short, for each example $i$, the contrastive learning aims to learn the embedding of {\tt [MASK]} and assign the appropriate label (entity type) to it by pulling semantically close examples with the same label together (positives) and pushing apart examples with a different label (negatives).

In case of having multiple examples in $P(i)$ and $N(i)$, we can rewrite the contrastive learning loss function as:
\begin{equation} \label{sec:sol:eq3}
L_C = -\log \sum_{i^+ \in P(i)}^{} \frac{e^{sim(t_i, t_{i^+})/\tau}}{\sum_{a \in A(i)}^{}e^{sim(t_i, t_{i^-})/\tau}}
\end{equation}
where $A(i)$ denotes a collection of all in-batch examples except $i$. Inspired by~\cite{gunel2021}, to fine-tune $\mathcal{L}$'s parameters and update the soft prompt parameters, we define the overall loss function $L$ as the weighted average of  $L_C$ and $L_S$, where $L_S$ is a cross-entropy to update the soft prompt parameters:
\begin{equation}
    \label{sol:cross}
    L_S = - \sum_{i \in \mathcal{B}} y_i \cdot \log \mathcal{L}(y'_i|\mathcal{T}(i))\\
\end{equation}
\begin{equation}
    L = \lambda L_C + (1-\lambda) L_SP
\end{equation}
where $\lambda$ is a scalar weighting hyperparameter that we tune, and $y_i$ and $y'_i$ are the correct and the predicted labels of $i$, respectively.

During inference, we first transform all test instances into the format of the soft-hard prompt template and then take the predicted label embedding $t_j$ of a test example $j$ to generate the label (entity type) directly by comparing $t_j$ to the k-nearest examples to $t_j$ in the training set.

\section{Evaluation} \label{experiments}
In this section, we conduct extensive experiments in standard and low-resource settings to evaluate \jina and its effectiveness in the few-shot NER settings.

\subsection{Datasets and Baselines} \label{experiments:datasets}
As a rich NER dataset, following~\cite{chen2021lightner} and~\cite{cui2021template}, we used the English version of CoNLL03~\cite{tjong-kim-2003-conll} that includes four features (columns) for each sample: {\tt id}, {\tt tokens}, {\tt pos\_tags}, {\tt chunk\_tags} and {\tt ner\_tags}. The feature {\tt tokens} represents the input sentence, and {\tt ner\_tags} specifies the type of mentioned entities in the input sentence, which contains four types of named entities: {\tt LOCATION}, {\tt PERSON}, {\tt ORGANIZATION}, and {\tt MISCELLANEOUS}. As low-resource datasets, we employed three datasets: (1) MIT Restaurant Review~\cite{Liu2013mit}, (2) MIT Movie Review~\cite{Liu2013mit}, and (3) Airline Travel Information Systems (ATIS)~\cite{hakkani2016multi}. To evaluate the few-shot performance on NER datasets, we randomly sampled $K$ instances per entity type from each low-resource dataset by setting $K$ to 10, 20, 50, 100, 200, and 500. We then reported the average performance of five randomly sampled data splits to avoid dramatic changes for different data splits.

In our experiments, we compare \jina with the following NER methods as baselines:
\begin{itemize}
    \item Sequence Labeling BERT/BART~\cite{devlin2018bert}: Traditional sequence labeling methods where the pre-trained BERT and BART~\cite{lewis2019bart} models are employed to generate word sequence representations. A label-specific classifier is trained on the top of PLM to map the generated representations to entity types (labels).
    \item LUKE~\cite{yamada2020luke}: A transformer-based model with an entity-aware self-attention layer that generates contextualized word representations. Then, the pre-trained model is fine-tuned using entity typing downstream task and linear classifiers to predict the type of an entity in the given sentence.
    \item BART-NER~\cite{yan-etal-2021-unified-generative}: A generative seq2seq method that converts the NER task into a unified sequence generation problem.
    \item TemplateNER~\cite{cui2021template} and LightNER~\cite{chen2021lightner}: Prompt-based models that use BART~\cite{lewis2019bart}.
\end{itemize}

We conducted the experiments on a Tesla T4 GPU with 32 cores and 576 GB of RAM. We used RoBERTa~\cite{Roberta2019}, provided by Hugging Face\footnote{https://huggingface.co/models}, as the PLM $\mathcal{L}$ in our implementation. We set $\tau$ = 2, $\lambda = 0.5 $, and trained the model using Adam optimizer~\cite{kingma2014adam} with a learning rate $5e-3$ and a batch size of 32.

\subsection{Standard NER Setting} \label{experiments:standardNER}
We first evaluated \jina using the CoNLL03~\cite{tjong-kim-2003-conll}, a rich NER dataset. Table~\ref{experiments:table:supervised} compares the results of \jina and the baselines. As shown in Table~\ref{experiments:table:supervised}, although we developed \jina for few-shot NER, it performs competitively in a rich-resource setting, showing the remarkable ability of our technique to identify the entities and their types in an input text.

\begin{table}[t]
    \centering
    \vspace{+1em}
    \caption{\footnotesize Model performance on the CoNLL03 dataset. '$\dagger$' shows the reported results with $BERT_{large}$~\cite{devlin2018bert} since the result of the original publication is not achieved with the current version of the library (See the discussion at~\cite{google2019bert} and the reported results at~\cite{Akbik2019NER}).}
    \vspace{+.5em}
    \footnotesize
    \scalebox{0.98}{
    \begin{tabular}{l|c|c|c}
    \hline
        \textbf{Traditional Models} & \textbf{$Precision$} & \textbf{$Recall$} & \textbf{$F1$}\\
    \hline
        Sequence labeling BERT$\dagger$ & 91.93 & 91.54 & 92.8 \\
        Sequence labeling BART & 89.60 & 91.63 & 90.60\\
        LUKE~\cite{yamada2020luke} & - & - & 94.30 \\
        BART-NER~\cite{yan-etal-2021-unified-generative} & 92.61 & 93.87 & 93.24\\
    \hline
        \textbf{Few-shot Friendly Models} & \textbf{$Precision$}& \textbf{$Recall$}& \textbf{$F1$} \\
    \hline    
        TemplateNER~\cite{cui2021template} & 90.51 & 93.34 & 91.90 \\
        LightNER~\cite{chen2021lightner} & 92.39 & 93.48 & 92.93 \\
        \jina & 91.04 & 93.44 & 92.22 \\
    \hline
    \end{tabular}
    }
    \label{experiments:table:supervised}
\end{table}

 \begin{figure*}[t]
\centering
\subfigure[MIT Movies dataset]{
    \includegraphics[width=0.3\linewidth]{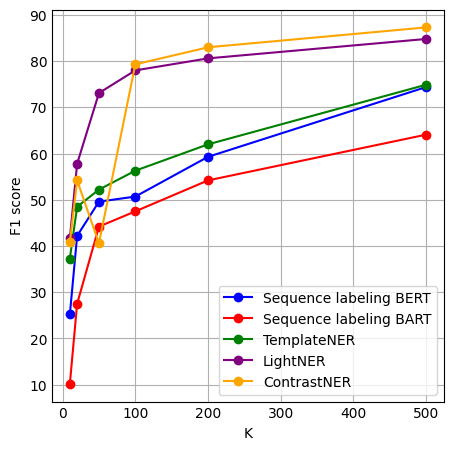}
    \label{fig:MITMovies}
}
\subfigure[MIT Restaurants dataset]{
    \includegraphics[width=0.3\linewidth]{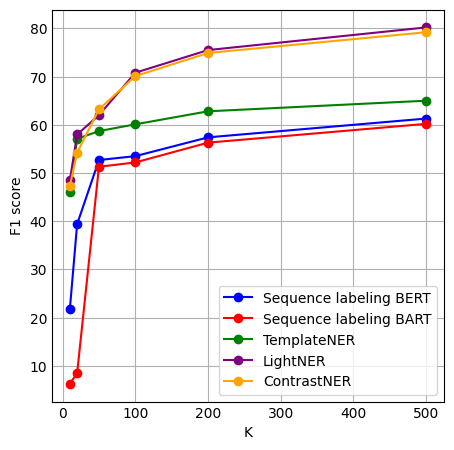}
    \label{fig:MITRestaurants}
}
\subfigure[ATIS dataset]{
    \includegraphics[width=0.3\linewidth]{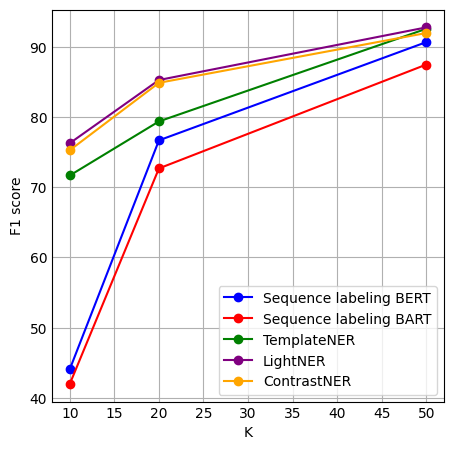}
    \label{fig:ATIS}
}
\caption[Optional caption for list of figures]{Model performance (F1 score) in the cross-domain low-resource settings when the model was trained on the same domain.}
\label{experiments:fig:withoutsource}
\end{figure*}

\subsection{In-domain Few-shot NER Setting} \label{experiments:in-domain}
Following~\cite{cui2021template}, we constructed a few-shot learning scenario on the CoNLL03 dataset, where the number of training samples for specific categories is limited by downsampling. Particularly, we considered {\tt PERSON} and {\tt ORGANIZATION} as the rich-resource entities and {\tt LOCATION} and {\tt MISCELLANEOUS} as the low-resource entities. The few-shot CoNLL03 training dataset contains 4237 training samples including 3836 {\tt PERSON}, 1924 {\tt ORGANIZATION}, 100 {\tt MISCELLANEOUS}, and 100 {\tt LOCATION}. Table~\ref{experiments:table:conll_few} indicates that \jina outperforms TemplateNER~\cite{cui2021template} and LightNER~\cite{chen2021lightner} by 3.68 and 0.3 $F1$ score, respectively. Moreover, \jina achieves 73.98 and 75.13 $F1$ scores in few-shot {\tt LOCATION} and {\tt MISCELLANEOUS}, which is highly competitive with the best-reported result. The illustrated performance proves the effectiveness of our approach in in-domain few-shot NER.
\begin{table}[t]
    \centering
    \caption{\footnotesize In-domain Few-shot performance on the CoNLL03. * indicates it is a few-shot entity type.}
    \vspace{+.5em}
    \footnotesize
    \scalebox{0.63}{
    \small
    \begin{tabular}{c|c|c|c|c|c}
         \hline
        Models & {\tt PERSON} & {\tt ORGANIZATION} & {\tt LOCATION}* & {\tt MISCELLANEOUS}* & Overall \\
        \hline
        Sequence labeling BERT & 76.25 & 75.68 & 60.72 & 59.39 & 68.02 \\
        Sequence labeling BART & 75.71 & 73.59 & 58.73 & 56.6 & 66.15 \\
        TemplateNER & 84.49 & 72.61 & 71.98 & 73.37 & 75.59 \\
        LightNER & 90.96 & \textbf{76.88} & \textbf{81.57} & 52.08 & 78.97 \\
        \jina & \textbf{92.19} & 75.79 & 73.98 & \textbf{75.13} & \textbf{79.27}\\
         \hline
\end{tabular}
}
\label{experiments:table:conll_few}
\end{table}

\subsection{Cross-Domain Few-Shot Setting}\label{experiments:cross-domain}
 Finally, we evaluated the model performance in scenarios where the class label sets and textual sentences vary from the source domain and only limited labeled data are available for training. Specifically, we randomly sampled a specific number of instances per entity type from the training set as the training data in the target domain to simulate the cross-domain low-resource data scenarios. We first considered direct training on the target domain from scratch without available source domain data. Figure~\ref{experiments:fig:withoutsource} depicts the results of training models directly on target domains and evaluation on the same domain. According to the results, compared to sequence labeling with BERT, BART, and TemplateNER, \jina's results appear more consistent. \jina outperforms these methods, suggesting it can exploit few-shot data better. For example, we achieved an $F1$ score of 70.6 in the 50-shot setting, which is higher than the results of sequence labeling with BERT and BART, and TemplateNER in the 200-shot setting. 

\begin{figure*}[t]
\centering
\subfigure[MIT Movies dataset]{
    \includegraphics[width=0.3\linewidth]{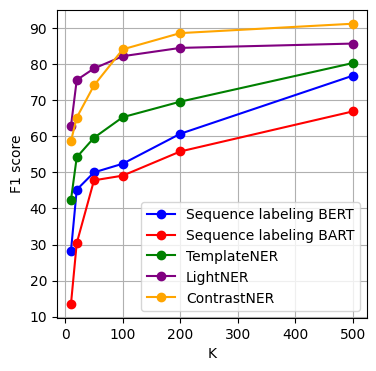}
    \label{fig:MITMovies}
}
\subfigure[MIT Restaurants dataset]{
    \includegraphics[width=0.3\linewidth]{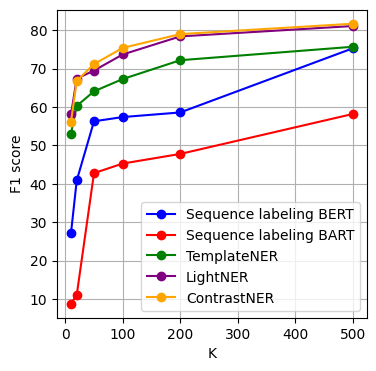}
    \label{fig:MITRestaurants}
}
\subfigure[ATIS dataset]{
    \includegraphics[width=0.3\linewidth]{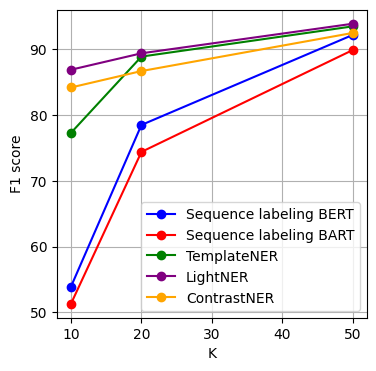}
    \label{fig:ATIS}
}
\caption[Optional caption for list of figures]{Model performance (F1 score) in the cross-domain low-resource settings when the model was trained on the CoNLL03 source domain and then evaluated on the target domains.}
\label{experiments:fig:withsource}
\end{figure*}

We then investigated the amount of knowledge that can be transferred from training on ConLL03 dataset. In this setting, we trained the model on the news domain (ConLL03) and then tested it on different domains. Figure~\ref{experiments:fig:withsource} shows the results of training models on the CoNLL03 dataset as a generic domain and its evaluations on other target domains. As can be seen, prompt-based methods outperform the traditional sequence labeling methods regardless of how much training data is provided. Among the prompt-based approaches, \jina indicates the best performance overall. Compared to LightNER, the best-performing NER framework among all state-of-the-art models, it can be witnessed that \jina can compete with this framework in different few-shot settings. At the same time, our approach does not require prompt engineering and verbalizer design. In particular, \jina outperforms LightNER in scenarios with more than 100 examples for each entity label. In other words, the contrastive loss is effective after a certain threshold of training data is reached. 

\subsection{Effectiveness of Soft-hard Prompt} \label{experiments:effectivenessPrompt}
\begin{figure}[t]
    \centering
    \resizebox{.3\textwidth}{!}{
    \includegraphics{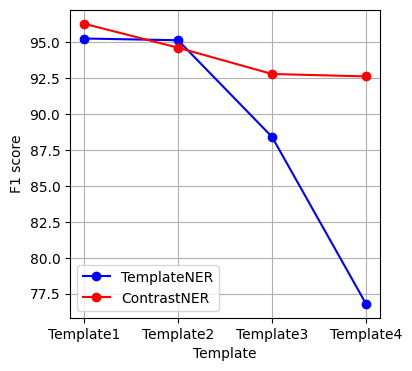}}
    \caption{\footnotesize Model performance ($F1$ score) on the CoNLL03 development set using different discrete prompt templates. '$\dagger$' shows the reported results in the original work.}
    \label{experiments:fig:prompt}
\end{figure}

In \jina, we applied the soft-hard prompt tuning technique that combines the soft and hard prompts. Their automatic search in a mixed space of continuous and discrete prompts eliminates the manual effort of prompt engineering. To investigate the effectiveness of our approach, we employed four different hard prompt templates, which are manually designed and used in TemplateNER~\cite{cui2021template} on the ConLL03~\cite{tjong-kim-2003-conll} development set (Table~\ref{experiments:table:discretePrompt}). Figure~\ref{experiments:fig:prompt} represents how selecting various discrete templates affects the performance of \jina and TemplateNER based on $F1$ score. As can be observed, our model produces more stable results, and unlike TemplateNER, the discrete template in \jina does not significantly influence the final performance. Since the soft tokens are adjusted during training, there is a slight variation in the model's performance when the discrete prompt template is changed, demonstrating the efficiency of using soft-hard prompts.

\begin{table}[hbt!]
    \centering
    \caption{\footnotesize Different discrete prompts applied on CoNLL03.}
    \vspace{+0.5 em}
    \footnotesize
    \scalebox{0.8}{
    \small
    \begin{tabular}{c|c}
        \hline
         & Discrete Template \\
         \hline
         Template1 & {\tt <candidate\_entity>} is a {\tt <entity\_type>} entity. \\
         \hline
         Template2 & The entity type of {\tt <candidate\_entity>} is {\tt <entity\_type>}. \\
         \hline
         Template3 & {\tt <candidate\_entity>} belongs to {\tt entity\_type>} category. \\
         \hline
         Template4 & {\tt <candidate\_entity>} should be tagged as {\tt <entity\_type>}. \\
         \hline
    \end{tabular}
}
    \label{experiments:table:discretePrompt}
\end{table}

\subsection{Discussion} \label{experiments:discussion}
This study investigated a prompt-based method in the few-shot NER problem. The results generally indicate that prompt-based NER approaches outperform the traditional NER frameworks in few-shot settings where only limited data is provided to train the model. Although prompt-based methods are specifically designed for low-resource scenarios, they also perform competitively with traditional methods in rich-resource scenarios. Determining how well previous prompt-based NER systems perform depends on discovering the best-performing prompt template (prompt engineering) or encountering the best label word space and their mapping to actual class labels (verbalizer engineering). At the same time, \jina exhibits more consistent performance with various discrete prompt templates and eliminates the need to determine the best-performing prompt template and optimal verbalizer manually. Despite this elimination, according to the results, it is relatively straightforward that \jina outperforms baselines in the in-domain few-shot scenario. Moreover, \jina indicates competitive performance in cross-domain scenarios. Due to the contrastive approach, \jina can likely beat the baselines after reaching a certain threshold of training data. It should be noted that this study is primarily concerned with removing the need for manual prompt and verbalizer engineering in the few-shot NER problem.

\section{Conclusion and Future Work} \label{conclusion}
This paper presents \jina, a prompt-based learning framework for few-shot NER without manual prompt engineering and design of verbalizers using RoBERTa~\cite{Roberta2019} as the backbone model. We present soft-hard prompt tuning for automatic prompt search in a mixed space of continuous and discrete prompts to avoid manual prompt engineering to find the best-performing prompt template. We also employ contrastive learning-based loss to unify learning the soft-hard prompt and predicting the entity type without manually designing a verbalizer. Our experiment results show that \jina indicates a competitive performance on both rich-resource and few-shot NER. In the future, we plan to extend \jina to the relation extraction task to develop a unified prompt-based learning framework for information extraction. Moreover, we will apply \jina on actual datasets from the \DataCloud project, which is a project on defining and managing Big Data pipelines in different applications including digital health systems, autonomous live sports content, and manufacturing analytics. We are going  to extract and store structured data from the definitions of Big Data pipelines defined in the form of unstructured natural language.

\bibliographystyle{IEEEtran} 
\bibliography{main.bib}

\end{document}